\documentclass[10pt,journal,letterpaper,compsoc]{IEEEtran}
\IEEEoverridecommandlockouts

\usepackage{float}
\usepackage[font=small,labelfont=bf]{caption}
\usepackage{subcaption}
\usepackage{subscript}
\usepackage[version=3]{mhchem}
\usepackage{setspace}
\usepackage{multirow}
\usepackage{amsfonts}
\usepackage{balance}
\newcommand\Tstrut{\rule{0pt}{2.6ex}}         
\newcommand\Bstrut{\rule[-0.9ex]{0pt}{0pt}}   

\def\BibTeX{{\rm B\kern-.05em{\sc i\kern-.025em b}\kern-.08em
    T\kern-.1667em\lower.7ex\hbox{E}\kern-.125emX}}

\usepackage{graphicx}
\graphicspath{{./}}
\DeclareGraphicsExtensions{.eps}

\begin{document}

\title{The Faults in Our $\pi^*$s:
	Security Issues and Open Challenges in Deep Reinforcement Learning
}
\author{
\IEEEauthorblockN{Vahid Behzadan}
\IEEEauthorblockA{\textit{Dept. of Computer Science} \\
	\textit{Kansas State University}\\
	Manhattan, KS, USA \\
	behzadan@ksu.edu}

\and
\IEEEauthorblockN{Arslan Munir}
\IEEEauthorblockA{\textit{Dept. of Computer Science} \\
\textit{Kansas State University}\\
Manhattan, KS, USA \\
amunir@ksu.edu} \\
}

\author
  {
  \IEEEauthorblockN
    {
    Vahid Behzadan and
    Arslan Munir\\
    }
    \IEEEauthorblockA
    {
    Department of Computer Science, Kansas State University \\
    Email:                                        %
	behzadan@ksu.edu,
    amunir@ksu.edu
    \vspace{-10 mm}
    }
  }

\maketitle

\begin{abstract}
	Since the inception of Deep Reinforcement Learning (DRL) algorithms, there has been a growing interest in both research and industrial communities in the promising potentials of this paradigm. The list of current and envisioned applications of deep RL ranges from autonomous navigation and robotics to control applications in the critical infrastructure, air traffic control, defense technologies, and cybersecurity. While the landscape of opportunities and the advantages of deep RL algorithms are justifiably vast, the security risks and issues in such algorithms remain largely unexplored. To facilitate and motivate further research on these critical challenges, this paper presents a foundational treatment of the security problem in DRL. We formulate the security requirements of DRL, and provide a high-level threat model through the classification and identification of vulnerabilities, attack vectors, and adversarial capabilities. Furthermore, we present a review of current literature on security of deep RL from both offensive and defensive perspectives. Lastly, we enumerate critical research venues and open problems in mitigation and prevention of intentional attacks against deep RL as a roadmap for further research in this area.
\end{abstract}


\vspace{-5 mm}
\section{Introduction}
\label{sec:introduction}
\vspace{-2 mm}
Since the inception of Artificial Intelligence (AI), a core objective of this science has been to pursue the creation of agents that autonomously learn skills and desired behaviors from self-guided interactions with the environment. This pursuit is greatly inspired by a biological parallel prevalent in natural learners such as humans, who learn from experience to repeat those behaviors that seem to result in higher rewards \cite{glimcher2011understanding}. Attempts to utilize this mechanism in the engineering of behaviors in natural agents are not new, as documented in psychological literature under paradigm of \emph{conditioning} \cite{dayan2002reward}. Adopting the grounds developed for natural agents, machine learning research has developed the framework of \emph{Reinforcement Learning} (RL) for artificial agents.

While the basic formulation and algorithms for RL were introduced decades back \cite{sutton1998reinforcement}, the feasibility and applicability of this framework were mostly severely constrained to simple problems and environments due to the computational requirements and the need for manual feature engineering \cite{szepesvari2010algorithms}. This was greatly overturned with the advancements in computing (e.g., GPUs), as well as the integration of deep learning with RL \cite{arulkumaran2017brief}. The latter provides RL with the powerful feature learning capabilities of deep neural networks, thus enabling the end-to-end learning of complex skills from raw inputs in complex environments. This new framework, known as \emph{Deep RL} (DRL), has led to many promising results in various settings, such as learning to play video games \cite{mnih2015human}, beating the human champion at the game of Go \cite{silver2016mastering}, and autonomous navigation \cite{peng2017deeploco}.

In recent years, the versatility and performance of DRL has motivated extensive research on its application to many domains, ranging from driverless vehicles \cite{fayjie2018driverless} and robotic manipulation \cite{gu2017deep} to healthcare \cite{raghu2017continuous}, algorithmic trading \cite{deng2017deep}, and automated control of smart grids \cite{franccois2016deep}, transportation systems \cite{madu2017urban} and air traffic management \cite{brittain2018towards}. In some domains, such as robotics \cite{eppner2016lessons} and portfolio management \cite{eppner2016lessons}, applications of DRL are grown beyond academic interest into deployed solutions.

With such rampant adoption of DRL into commercial and critical systems, ensuring the security of these applications is of growing importance. As a major component of cyber-physical and financial systems, DRL controllers present an attractive attack surface to adversaries aiming to manipulate or otherwise compromise the functions of dependent systems. As detailed in Section \ref{sec:threat}, the security problem is different from those that the area of safe RL \cite{garcia2015comprehensive} are concerned with. The objective of safe RL is to ensure that the agent does not learn to behave in ways that are not in compliance with some pre-defined criteria. For instance, one criteria for safety in RL-based agent trained for autonomous navigation is to learn a policy that avoids collisions. The security problem is however concerned with settings in which an adversarial element intentionally seeks to compromise the natural operation of the system for malicious purposes. In the example of autonomous navigation, an agent with a provably safe RL policy can be forced into collisions via means such as the intentional perturbation of sensory input \cite{behzadan2018adversarial}, as detailed in the following sections. Hence, while there are overlapping areas between RL safety and security, a different approach is required to study the security problem.

While the security of supervised and unsupervised machine learning systems has enjoyed extensive attention from the research community \cite{papernot2018sok}\cite{papernot2016towards}\cite{biggio2018wild}, the work on vulnerabilities and security of DRL is sparse and sporadic. Since the reports by Behzadan \& Munir \cite{behzadan2017vulnerability} and Huang et al. \cite{huang2017adversarial} in early 2017, few papers have studied this problem with focus on narrow aspects of security in DRL. Thus, there remains a need for a holistic formulation of the security issues in DRL, as well as a roadmap for further researcher.

This paper aims to fill this gap by providing a comprehensive analysis of the security threats in DRL from a technical perspective. Furthermore, we study the state of the art in current literature and identify major venues of research that remain largely unexplored. Accordingly, the main contributions of this paper are:
\vspace{-2 mm}
\begin{itemize}
	\item Formulation of the security problem in DRL;
	\item Proposal of a generic threat model in terms of attack surface in DRL, adversarial capabilities, and attack objectives;
	\item Formulation and analysis of attacks at both training and inference phases;
	\item Review of current literature on attacks and defenses in DRL
	\item Enumeration of critical open research problems and challenges in ensuring the security of DRL
\end{itemize}

\vspace{-2 mm}
The remainder of this paper is organized as follows: Section \ref{sec:background} reviews the fundamentals of RL and DRL, and provides an overview of adversarial machine learning. Section \ref{sec:threat} formalizes the security problem in RL, and presents threat models for adversarial attacks against DRL. Section \ref{sec:state} reviews the current state of the art in the types and mechanisms for attacks at training and inference phases of DRL, the defenses against such attacks, simulation benchmarks, and evaluation metrics. In Section \ref{sec:open} multiple critical venues requiring further research are identified, and Section \ref{sec:conclusions} concludes the paper with a summary of the findings.
%

\vspace{-3.5 mm}
\section{Background}
\label{sec:background}
\vspace{-0.5 mm}
\subsection{Deep Reinforcement Learning}
\label{sec:DRL}
\vspace{-1 mm}
Reinforcement learning is concerned with agents that interact with an environment and exploit their experiences to optimize a decision-making policy. The generic RL problem can be formally modeled as a Markov Decision Process (MDP), described by the tuple $MDP = (S, A, R, P)$, where $S$ is the set of reachable states in the process, $A$ is the set of available actions, $R$ is the mapping of transitions to the immediate reward, and $P$ represents the transition probabilities (i.e., dynamics), which are initially unknown to RL agents. At any given time-step $t$, the MDP is at a state $s_t\in S$. The RL agent's choice of action at time $t$, $a_t \in A$ causes a transition from $s_t$ to a state $s_{t+1}$ according to the transition probability $P_{s_t , s_{t+1}}^{a_t}$. The agent receives a reward $r_{t+1}$ for choosing the action $a_t$ at state $s_t$. Interactions of the agent with MDP are determined by the policy $\pi$. When such interactions are deterministic, the policy $\pi: S\rightarrow A$ is a mapping between the states and their corresponding actions. A stochastic policy $\pi(s)$ represents the probability distribution of implementing any action $a\in A$ at state $s$. The goal of RL is to learn a policy that maximizes the expected discounted return $E[R_t]$, where $R_t = \sum_{k=0}^{\infty}\gamma^k r_{t+k}$; with $r_t$ denoting the instantaneous reward received at time $t$, and $\gamma$ is a discount factor $\gamma\in [0,1]$. The value of a state $s_t$ is defined as the expected discounted return from $s_t$ following a policy $\pi$, that is, $V^{\pi}(s_t) = E[R_t|s_t, \pi]$. The action-value (Q-value) $Q^{\pi}(s_t, a_t) = E[R_t|s_t,a_t, \pi]$ is the value of state $s_t$ after using action $a_t$ and following a policy $\pi$ thereafter.

There are three main approaches to solving RL problems according to their optimization objective: methods that are based on \emph{value functions}, those that are based on \emph{policy search}, and a hybrid of both in \emph{actor-critic} configurations. The details of each approach is presented as follows:

\vspace{-2.5 mm}
\subsubsection{Value Iteration and Deep Q-Network}
\vspace{-1 mm}
Value iteration refers to a class of algorithms for RL that optimize a value function (e.g., V(.) or Q(.,.)) to extract the optimal policy from it. As an instance of value iteration algorithms, \emph{Q-Learning} aims to maximize for the action-value function $Q$ through the iterative formulation of Eq.~(\ref{bellman}):
\begin{eqnarray} \label{bellman}
Q(s,a) = R(s, a) + \gamma max_{a'}(Q(s',a'))
\end{eqnarray}

Where $s'$ is the state that emerges as a result of action $a$, and $a'$ is a possible action in state $s'$. The optimal $Q$ value given a policy $\pi$ is hence defined as: $Q^\ast (s, a) = max_{\pi} Q^{\pi} (s, a)$, and the optimal policy is given by $\pi^\ast(s) = \arg\max_a Q(s,a)$.

The Q-learning method estimates the optimal action policies by using the Bellman formulation $Q_{i+1} (s,a) = \mathbf{E}[R + \gamma \max_a Q_i]$ as the iterative update of a value iteration technique. Practical implementation of Q-learning is commonly based on function approximation of the parametrized Q-function $Q(s,a; \theta) \approx Q^\ast (s,a)$. A common technique for approximating the parametrized non-linear Q-function is via neural network models whose weights correspond to the parameter vector $\theta$. Such neural networks, commonly referred to as Q-networks, are trained such that at every iteration $i$, the following loss function is minimized:
\begin{eqnarray}\label{eq:update}
L_i(\theta_i) = \mathbf{E}_{s, a\sim \rho(.)} [(y_i - Q(s,a,;\theta_i))^2]
\end{eqnarray}

where $y_i = \mathbf{E}[R + \gamma \max_{a'}Q(s',a';\theta_{i-1}) | s,a]$, and $\rho(s,a)$ is a probability distribution over states $s$ and actions $a$. This optimization problem is typically solved using computationally efficient techniques such as Stochastic Gradient Descent (SGD).

Classical Q-networks introduce a number of major problems in the Q-learning process. First, the sequential processing of consecutive observations breaks the \emph{iid} (Independent and Identically Distributed) requirement of training data as successive samples are correlated. Furthermore, slight changes to Q-values leads to rapid changes in the policy estimated by Q-network, thus enabling policy oscillations. Also, since the scale of rewards and Q-values are unknown, the gradients of Q-networks can be sufficiently large to render the backpropagation process unstable.

A Deep Q-Network (DQN) \cite{mnih2015human} is a training algorithm designed to resolve these problems. To overcome the issue of correlation between consecutive observations, DQN employs a technique called \emph{experience replay}: instead of training on successive observations, experience replay samples a random batch of previous observations stored in the replay memory to train on. As a result, the correlation between successive training samples is broken and the \emph{iid} setting is re-established. In order to avoid oscillations, DQN fixes the parameters of a network $\hat{Q}$, which represents the optimization target $y_i$. These parameters are then updated at regular intervals by adopting the current weights of the Q-network. The issue of unstability in backpropagation is also solved in DQN by normalizing the reward values to the range $[-1,+1]$, thus preventing Q-values from becoming too large.

Mnih et al. \cite{mnih2015human} demonstrate the application of this new Q-network technique to end-to-end learning of Q values in playing Atari games based on observations of pixel values in the game environtment. To capture the movements in the game environment, Mnih et al. use stacks of four consecutive image frames as the input to the network. To train the network, a random batch is sampled from the previous observation tuples $(s_t, a_t, r_t, s_{t+1})$, where $r_t$ denotes the reward at time $t$. Each observation is then processed by two layers of convolutional neural networks to learn the features of input images, which are then employed by feed-forward layers to approximate the Q-function. The target network $\hat{Q}$, with parameters $\theta^{-}$, is synchronized with the parameters of the original $Q$ network at fixed periods intervals. i.e., at every $i$th iteration,  $\theta^-_{t} = \theta_t$, and is kept fixed until the next synchronization. The target value for optimization of DQN thus becomes:
\vspace{-1.5 mm}
\begin{eqnarray}
y'_t \equiv r_{t+1} + \gamma max_{a'} \hat{Q}(s_{t+1}, a'; \theta^-)
\end{eqnarray}

Accordingly, the training process can be stated as:
\vspace{-1 mm}
\begin{eqnarray}\label{SGD}
min_{a_t} (y'_t - Q(s_t, a_t, \theta))^2
\end{eqnarray}

As for the exploration mechanism, the original DQN employs the $\epsilon$-greedy technique, which monotonically decreases the probability of taking random actions as the training progresses \cite{sutton1998reinforcement}. However, recent literature proposes various alternatives such as parameter-space noise exploration \cite{fortunato2017noisy}. Although adding independent noise for exploration is usable in continuous control problems, more sophisticated strategies inject noise that is correlated over time (e.g., from stochastic processes) in order to better preserve momentum \cite{lillicrap2015continuous}.

In \cite{hasselt2010double}, Hasselt et al. demonstrate that the single estimator used in the update rule of Q-learning (provided in Eq~(\ref{eq:update})) overestimates the expected return, as it uses the maximum action value as an approximation of the maximum expected action value. As a solution, \cite{van2016deep} presents a generalization of the Double Q-learning algorithm \cite{hasselt2010double}, and develops Double DQN (DDQN) that separates action selection and action evaluation, i.e., one DQN is used to determine the maximizing action and a second one is used to estimate its value.

\vspace{-2 mm}
\subsubsection{Policy Search Methods}
\vspace{-1 mm}
The optimization objective of policy search methods is to directly find policies via either gradient-free or gradient-based methods. While gradient-free methods such as evolutionary methods have resulted in some success in this area, much of the recent developments are focused on policy gradient (i.e., gradient-based) methods \cite{arulkumaran2017brief}. In policy gradient methods, the policy is directly parametrized in the form $\pi(a|s;\theta)$, where $\pi$ is a probability distribution over actions $a$ when observing state $s$, as parameterized by $\theta$, which can be a neural network. The agent exercises this policy in the environment and collects experiences. Periodically, it uses the experience samples to update $\theta$ by estimating the gradient $\nabla_\theta E[R_t]$. Typically, the agent then discard these samples and repeats this process on new samples, optimizing the policy iteratively. Two of the most significant algorithms for solving policy gradient are TRPO and PPO, detailed below:

\textbf{Trust Region Policy Optimization (TRPO):}
TRPO has been shown to be relatively robust and applicable to domains with high-dimensional inputs \cite{schulman2015trust}. To achieve this, TRPO optimizes a surrogate objective function, specifically, it optimizes an (importance sampled) advantage estimate, constrained using a quadratic approximation of the Kullback-Leibler (KL) divergence. Whilst TRPO can be used as a pure policy gradient method with a simple baseline, later work by Schulman et al. \cite{schulman2015high} introduces Generalized Advantage Estimation (GAE), which proposes several, more advanced variance reduction baselines. The combination of TRPO and GAE remains one of the stateof-the-art RL techniques in continuous control. However, the constrained optimization of TRPO requires calculating second-order gradients, limiting its applicability.

\textbf{Proximal Policy Optimization (PPO):}
PPO \cite{schulman2017proximal} is a closely related algorithm that improves sample complexity (i.e., the number of samples required to learn an optimal policy) by increasing the training use of each sample. It maximizes a ``surrogate'' objective $E[\rho_t(\theta)A_t]$, where $\rho_t(\theta) = \pi(a_t|s_t;\theta)/\pi(a_t|s_t;\theta_{old})$ is the likelihood ratio of the recorded action between the updated and original policies. Unlike A3C, PPO performs multiple parameter updates using minibatches from each set of samples.

\vspace{-3 mm}
\subsubsection{Advantage Actor-Critic}
\vspace{-1 mm}
Actor-critic methods combine value functions with an explicit representation of the policy. In such methods, the actor represents a policy that leverages the feedback from the value function (i.e., critic) to optimize towards an optimal policy. In advantage actor-critic methods, the policy gradient is computed as $E[\nabla_\theta \log \pi(a_t|s_t;\theta)(R_t-V(s_t))]$. The agent estimates $V(s_t)$ from the data via a separate output from the same network used for $\pi$. $(R_t-V(s_t))$ estimates the advantage $A(s,a) = Q(s,a) - V(s)$. $R_t$ is computed using the discounted sum of as many future returns as are observed in a given batch, up to $r_{t_{max}}$, where $t_{max}$ is the horizon of interest to the agent (e.g., end of an episode). The estimator $V(s;\theta)$ is trained according to squared-error loss, simultaneously to $\pi$. Lastly, an entropy bonus is added to the gradient: $\nabla_\theta H(\pi(.|s_t;\theta))$, to promote exploration and discourage premature convergence.

Notable instances of Advantage Actor-Critic methods include:

The \emph{Deep Deterministic Policy Gradient (DDPG)} \cite{lillicrap2015continuous} is an improvement of the original DPG algorithm \cite{silver2014deterministic}, adding experience replay and target networks. Experience is collected into a buffer and updates to actor and critic models are computed using mini-batch updates with random samples from this buffer. Furthermore, a second set of target networks is maintained for use in computing the loss.

The \emph{Asynchronous Advantage Actor-Critic (A3C)} algorithm \cite{mnih2016asynchronous} is comprised of separate actor-learner threads that sample environment steps and update a centralized copy of the parameters asynchronously to each other.

On the other hand, the \emph{Advantage Actor-Critic (A2C)} algorithm \cite{wang2016learning} uses a single-threaded learner to sample from separate environment instances and collects all data into one minibatch to compute the gradient.

\vspace{-4 mm}
\subsection{Security of Machine Learning}
\label{sec:mlsec}
\vspace{-1 mm}
As data-driven systems, machine learning algorithms are known to be vulnerable to various types of adversarial actions. Such actions can be broadly classified as those affecting the training phase of the learning process, and those targeting the inference phase (i.e., test-time) \cite{papernot2018sok}.

In the training phase, adversaries may aim to influence the learning process by manipulating the training data. This type of attack is generally referred to as \emph{poisoning} \cite{biggio2012poisoning}. An example of poisoning attacks is the case of online spam classifiers, where an adversary can intentionally mislabel spam emails as benign to compromise the accuracy of the model retrained on new data \cite{nelson2010behavior}.

In the inference phase, the adversary may implement an \emph{evasion} attack by providing malicious input that induce incorrect inferences at the output of a machine learning model \cite{biggio2013evasion}. This type of malicious input is generally referred to as adversarial example \cite{szegedy2013intriguing}. A noteworthy property of adversarial examples in their transferability: an adversarial example crafted for a particular model may also affect other models with different architectures that are trained on datasets with similar distributions to that of the original model \cite{tramer2017space}.

According to the objective of adversaries, adversarial example attacks are generally classified into the following two categories:
\vspace{-2 mm}
\begin{enumerate}
	\item Misclassification attacks, which aim for generating examples that are classified incorrectly by the target network
	\item Targeted attacks, whose goal is to generate samples that the target misclassifies into an arbitrary class designated by the attacker.
\end{enumerate}

\vspace{-2 mm}
To generate such adversarial examples, several algorithms have been proposed, such as the Fast Gradient Sign Method (FGSM) by Goodfellow et al., \cite{goodfellow6572explaining}, and the Jacobian Saliency Map Algorithm (JSMA) approach by Papernot et al., \cite{papernot2016limitations}. A grounding assumption in many of the crafting algorithms is that the attacker has complete knowledge of the target neural networks such as its architecture, weights, and other hyperparameters. In response, Papernot et al. \cite{papernot2017practical} proposed the first blackbox approach to generating adversarial examples. This method exploits the transferability of adversarial examples: an adversarial example generated for a neural network classifier applies to most other neural network classifiers that perform the same classification task, regardless of their architecture, parameters, and even the distribution of training data. Accordingly, the approach of \cite{papernot2017practical} is based on generating a replica of the target network. To train this replica, the attacker creates and trains over a dataset from a mixture of samples obtained by observing target's interaction with the environment, and synthetically generated inputs and label pairs. Once trained, any of the algorithms that require knowledge of the target network for crafting adversarial examples can be applied to the replica. Due to the transferability of adversarial examples, the perturbed data points generated for the replica network can induce misclassifications in many of the other networks that perform the same task.

Another class of adversarial actions is comprised of those that aim to infer information about the internal parameters of the model or the training dataset \cite{papernot2018sok}. One instance of such attacks is that of model extraction \cite{tramer2016stealing}, in which the adversary estimates the parameters of a model from observations of its input-output data points. Besides compromising the confidentiality of models as intellectual properties, an adversary may utilize the extracted model in circumventing the difficulties of blackbox attacks \cite{papernot2017practical}. Other instances of attacks on confidentiality are training data extraction \cite{fredrikson2014privacy} and membership attacks \cite{shokri2017membership}, which aim to infer information about the training set and extract personally-identifiable information from the model.

While the current literature on machine learning security includes various proposals for mitigation of these attacks, majority of solutions provide ad hoc alleviation of the problem and are not generalizable to other classes of algorithms. For training-time attacks, notable solutions are those that aim to minimize the impact of outliers in models based on Principle Component Analysis (e.g., \cite{rubinstein2009antidote}) and Support Vector Machines (e.g., \cite{biggio2011support}). As for attacks targeting the inference phase, notable solutions include gradient masking (e.g., \cite{gu2014towards} and \cite{papernot2015distillation}) and injecting adversarial examples in the training dataset (e.g., \cite{goodfellow6572explaining}). Yet, all such defenses are shown to be weak against adaptive adversaries \cite{papernot2018sok}. For more in-depth studies on the state of security in machine learning, readers can refer to \cite{papernot2018sok} and \cite{biggio2018wild}.

\vspace{-3 mm}
\section{Threat Model}
\label{sec:threat}
\vspace{-1 mm}
As with other types of machine learning techniques, DRL is also prone to adversarial actions similar to those mentioned in Section \ref{sec:mlsec}. While many aspects of DRL are similar to other machine learning techniques such as deep learning classifiers, the inherent differences in their learning dynamics and applications gives rise to security issues that are specific to DRL. One such difference is in the nature of the training data: in general, supervised learning is trained on a dataset sampled from a fixed distribution. One consequence of this fact is that poisoning attacks cause a shift in the distribution, and hence detection and mitigation of such shifts has been the primary approach in defending against poisoning attacks. However, the exploration mechanism in DRL inevitably results in changes in the distribution of training samples. In other words, both the attack mechanism and any defensive approach against poisoning attacks on DRL are fundamentally different from those of supervised learning. Furthermore, DRL agents are typically tasked with solving sequential decision-making problems. In many cases, this task implies a delayed-reward mechanism. For example, when playing chess, one may not receive any reward for any of its actions until the game is over. This is fundamentally different from supervised learning, where the availability of labels enable straightforward calculation of accuracy and other performance metrics for the model. Also, even after the training phase, many DRL agents retain a level of randomness in their output actions (i.e., the action $a_t$ is not always derived from a deterministic policy). As a result, distinguishing adversarial attacks from benign actions derived from stochastic policies is not as straightforward as the case of supervised learning. Indeed, it can be shown that both supervised and unsupervised mechanisms can be reduced from the RL framework (i.e., can be formulated as instances of RL), but the inverse is not true \cite{barto2004reinforcement}. In other words, DRL inherits the fundamental security issues in supervised and unsupervised learning, as well as other issues that are unique to RL and DRL.

As detailed in Section \ref{sec:mlsec}, adversarial attacks against DRL agents aim to compromise the normal operating criteria of such agents. From the standpoint of cybersecurity, the compromise may be viewed as affecting one or more dimensions of the Confidentiality, Integrity, Availability (CIA) triad \cite{pfleeger2012analyzing}, as detailed below:

\emph{Confidentiality} of a DRL agent refers to the need for keeping the internal configurations of the agent from exposure to adversaries. These configurations include agent's reward function, the training mechanism (e.g., hyperparameters, algorithm), and the learned policy function. For instance, inference of the policy function can result in loss of proprietary assets. Furthermore, knowledge of the internal configurations may enable the adversary to launch further attacks with more precision and efficiency.

\emph{Integrity} in DRL is the ability to learn or enact policies in the manner intended by the designer. Adversaries may compromise the integrity of a DRL agent by forcing it to learn incorrect policies, or to perform actions other than those prescribed by a learned policy.

\emph{Availability} is the ability of a DRL agent to perform training or actions when needed. Adversarial compromise of this dimension may be in the form of denying convergence during DRL training, or preventing the agent from acting in response to changes in the environment.

The problem of DRL security may resemble those studied under safe RL \cite{garcia2015comprehensive}, but as mentioned in Section \ref{sec:introduction}, the presence of adversarial intention in the security problem gives rise to challenges that are beyond the scope of safe RL. Also, we must differentiate between DRL security and the area of adversarial RL \cite{uther1997adversarial}. The latter is concerned with multi-agent RL settings, in which agents aim to maximize their returns in competition with other agents. While some DRL security problems can be modeled as adversarial RL (e.g., \cite{behzadan2018adversarial}), this cannot be generalized as the adversary is not necessarily a learning agent.
%

\vspace{-3 mm}
\subsection{DRL Attack Surface}
\label{sec:surface}
\begin{figure}[t]
	\centering
	
	\includegraphics[width = \columnwidth]{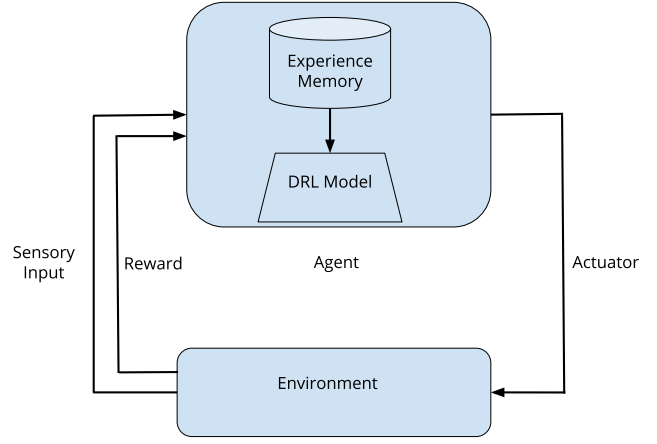}
	\vspace{-5 mm}
	\caption{Components of a DRL agent.}
	\label{fig:DRLComp}
	\vspace{-5 mm}	
\end{figure}

\vspace{-1 mm}
Figure \ref{fig:DRLComp} depicts the major components of a DRL agent in both training and inference phases. Each of these components can be targeted in adversarial attacks, as detailed below:

\emph{Environment}: Recall the MDP formulation of Section \ref{sec:DRL} - all aspects of the interaction between a DRL agent and the environment at time $t$ are captured by the tuple $(s_t, a_t, r_{t+1}, s_{t+1})$. It is observed that a fundamental input to both training and inference processes is the agent's observation of the environment's state $s_t$ and $s_{t+1}$. An adversary may perturb the environment and its configuration by some vector $\delta_t$ (i.e, $s'_t = s_t + \delta_t$ ) to manipulate the training or inference of the agent. For instance, Behzadan \& Munir \cite{behzadan2018adversarial} demonstrate that through sequential reconfiguration of obstacles on a road, an adversary can manipulate the trajectory of a DRL-based autonomous vehicle at test-time.

\emph{Sensors}: DRL agents observe the environment via their sensors, and the sensory observation $o_t$ of the state $s_t$ may contain noise $\alpha_{s_t}$, that is, $o_t = s_t + \alpha_{s_t}$. Adversarial manipulation of $\alpha_{s_t}$ through perturbing the sensor readings can compromise both the training and inference processes. An example is sequential blinding of the cameras in a DRL-based autonomous vehicle via lasers, which can lead to learning incorrect policies during training or denial of service at training time.

\emph{Reward Signal}: Manipulation of the reward signal $r_t$ produced by the environment in response to the actions of a DRL agent can greatly affect the training process. This case constitutes an instance of the \emph{corrupted reward channel} problem \cite{everitt2017reinforcement}, and can be formulated as follows: Given a true reward function $\dot{R}$ and a corruption function $C$, the observed reward function is defined as $\hat{R}: S\rightarrow \hat{R}$ as $\hat{R}(s):= C_s(\dot{R})$. The corrupted reward thus induces an observed MDP $(S, A, R, P, \hat{R})$, which may poison the experience memory with potentially corrupted observations of rewards. In the example of the autonomous vehicle, if the reward function depends on the distance of DRL agent to a destination as measured by GPS coordinates, spoofing of GPS signals by the adversary may result in incorrect reward signals, which can translate to incorrect navigation policies.

\emph{Experience Memory}: If the adversary can access and manipulate the experience memory of a DRL agent, it is possible to greatly influence the training process of the DRL agent. In the autonomous vehicle example, the agent can exploit physical or software vulnerabilities to access the replay memory during or in between training sessions and mount a poisoning attack.

\emph{Actuator}: DRL agents influence their environments by performing actions $a_t$ via actuators. If the adversary can manipulate the actuator, the actual action performed will be different than that chosen by the agent, and hence the observed experience is corrupted, which can translate into poisoning of the experience memory.

\vspace{-2 mm}
\subsection{Adversarial Capabilities}
\label{sec:capabilities}
\begin{figure}[!h]
	\centering
	\includegraphics[width = \columnwidth]{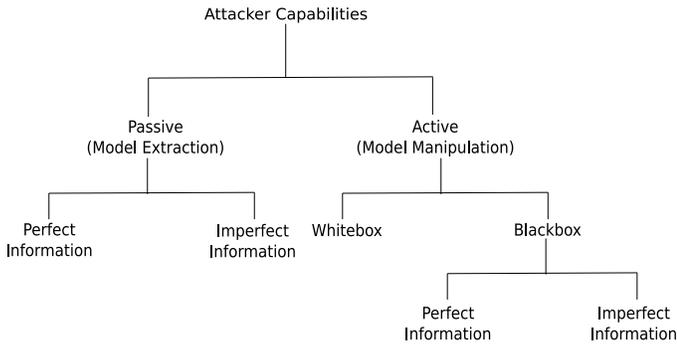}
	\vspace{-4 mm}
	\caption{Attacker capabilities.}
	\label{fig:ACDiagram}
	\vspace{-6 mm}
\end{figure}

\vspace{-1 mm}
Threat modeling of the adversary is comprised of two components, actions available to the adversary, and the information at his disposal. These capabilities, summarized in Figure~\ref{fig:ACDiagram}, define the adversarial constraints and determine the feasibility of attack vectors in the threat landscape of a DRL agent. The attacker may be able to launch attacks in either passive or active modes. Passive attacks do not disturb the normal functioning of the target, but aim to infer information about the parameters and configuration of the target's model via observation of state-inference pairs. On the other hand, active attacks aim to manipulate the output of the model.

The availability of a priori information to the attacker can be explored with respect to the type of information. Adversary's knowledge of the target model, its parameters, algorithm, and the reward function constitute the \emph{agent information}, while information about the dynamics and configuration of the environment comprise the \emph{environment information}. The adversary may have varying degrees of access to either type of information. For instance, adversary may have complete knowledge of the environment's dynamics, but may only have access to partial or noisy observations of changes in the state of the environment. Similarly, the set of actions available to the attacker may include perturbing the environment, the observation, or the reward signal, experience memory, or actuators in a DRL setting, constrained by a budget $B$ that determines the permitted extent of actions. For instance, in crafting visual adversarial examples to perturb observations, the adversary may be limited to a maximum of $B_p$ pixel modifications at each step. In the same example, the budget may also limit the frequency of perturbations to a maximum of one frame per every $B_f$ samples.

We explore the the different attack vectors available to adversaries according to the targeted learning phase (training or testing), as detailed in the following:

\vspace{-2 mm}
\subsubsection{Inference Phase}
Similar to test-time attacks discussed in Section \ref{sec:mlsec}, inference-time attacks against DRL do not tamper the learned policy and experience data of the target. Passive attacks at this phase aim for extraction of agent information, while active attacks at inference-time are similar to \emph{evasion attacks}, as both aim to manipulate the output of the agent.

Depending on the information available to the adversary, attacks can take various forms. If the adversary has access to the agent information at a level sufficient for precise crafting of malicious input (e.g., adversarial examples) to the agent, then the attack is considered \emph{whitebox}. Conversely, if such information is not available to the adversary, adversarial action is constrained to \emph{blackbox} attacks. Similarly, if the agent has access to a degree of knowledge of the environment that is sufficient for direct exploitation in adversarial actions, the attack is of the \emph{perfect information} type. For instance, knowledge of the dynamics of the Atari game Pong, as well as perfect observation of target's actions, enables the adversary to predict the future states and exploit this foresight in devising a sequence of perturbations to manipulate the DRL agent into performing poorly. On the other hand, if the adversary has complete knowledge of the game dynamics but can only have noisy or intermittent observability on the agent's actions, utilizing predictive control for adversarial actions may be rendered infeasible and the adversary will have to pursue attacks of type \emph{imperfect information}.

\vspace{-2 mm}
\subsubsection{Training Phase}
\vspace{-1 mm}
The objective of attacks targeting the training phase is to extract, manipulate, or disable the policy learning mechanism of the DRL agent. Similar to those under the inference phase, extraction attacks are generally passive, and aim to extract the model parameters or contents of experience memory. A simple instance of extraction attacks at training phase is direct access to the training software or memory. Other techniques may include inference of parameters based on observations of the training process. Analogous to training-time attacks against supervised learners \cite{biggio2011support}, manipulation or disruption of the training process can be achieved via two strategies: injection and modification. Injection refers to alteration of the experience memory via insertion of experience tuples that change the distribution of observations. A suitable technique for injection attacks is the manipulation of the environment itself. Modification attacks aim to change the contents of actual experiences. Adversarial example attacks against DRL at training time are a representative example of such attacks.

%

\vspace{-4 mm}
\section{State of the Art}
\label{sec:state}
\vspace{-1.5 mm}
In this section, we present an overview of the prominent literature on the security of DRL. While this body of work is still at its early stages, the research on security of DRL has produced notable results which can form the foundation for further advancements in this domain. Majority of the relevant papers can be classified under either offensive or defensive proposals. Furthermore, with the exception of two papers, offensive proposals are generally focused on either inference-time attacks or training-time attacks. Accordingly, we adopt the same scheme in our exploration of these literature. A summary of papers on attack mechanisms is presented in Table \ref{table:Attacks}, and the summary of literature on defensive techniques is provided in Table \ref{table:defenses}. It must be noted that while this review is prepared with the aim of producing a comprehensive survey of relevant literature, due to the fast rate of publications in machine learning, some recent yet notable papers may be missing from this review.
\begin{table*}[!thb]
	\caption{Summary of literature on attack mechanisms.}
	\vspace{-2 mm}
	\small
	\centering
	\begin{tabular}{ |c|c|c|c|c| }
		\cline{1-5}
		\hline
		\multirow{2}{*}{\textbf{Phase}} & \multirow{2}{*}{\textbf{Mode}} & \multicolumn{3}{c|}{\textbf{Attack Surface}}\Tstrut\Bstrut\\
		
		\cline{3-5}
		& \multicolumn{1}{|c|}{} & \textbf{Observation} & \textbf{State} & \textbf{Reward}\Tstrut\Bstrut\\ \cline{1-5}
		\multirow{5}{*}{\textbf{Test-Time}} &\multirow{3}{*}{Whitebox} & & & \\
		& & Behzadan \& Munir \cite{behzadan2017vulnerability}, Huang et al. \cite{huang2017adversarial}, & &\\ & & Patthanaik et al. \cite{pattanaik2018robust},  Lin et al. \cite{lin2017tactics}, & Han et al \cite{han2018reinforcement} , Clark et al. \cite{clark2018malicious} & Han et al. \cite{han2018reinforcement}\\
		& & Tretschk et al. \cite{tretschk2018sequential} & & \Tstrut\Bstrut\\ \cline{2-5}
		& & & & \\ & \multirow{1}{*}{Blackbox} & Behzadan \& Munir \cite{behzadan2017vulnerability}, Huang et al. \cite{huang2017adversarial} & Behzadan \& Munir \cite{behzadan2018adversarial} & Han et al. \cite{han2018reinforcement} \\ & & & & \\ \cline{2-5}
		\hline
		\multirow{5}{*}{\textbf{Training-Time}} & & & & \\ & Whitebox & Kos \& Song \cite{kos2017delving} & --- & Han et al. \cite{han2018reinforcement}\\ & & & & \\ \cline{2-5}
		& & & & \\ & Blackbox & Behzadan \& Munir \cite{behzadan2017vulnerability} & --- & --- \\ & & & & \\ \cline{3-5}
		\hline
		
	\end{tabular}
	\label{table:Attacks}
	\vspace{-5 mm}
\end{table*}

\vspace{-3.5 mm}
\subsection{Test-time Attacks}
\label{sec:testtime}
\vspace{-1 mm}
Majority focused on perturbing the observations of the agent.

Behzadan \& Munir \cite{behzadan2017vulnerability} published the first report on test-time vulnerabilities of DRL. Based on the fact that DQN policies and deep classifiers are essentially of the same structure and function, therefore DQN policies must also be vulnerable to adversarial examples. Accordingly, their paper tests the applicability of adversarial examples crafted with FGSM and JSMA versus DQN policies under whitebox settings, and demonstrate that such policies are also vulnerable to test-time manipulation using adversarial examples. In whitebox settings, the adversary performs a Man in The Middle (MITM) attack. He observes the state of the environment, and with complete and perfect knowledge of the target’s policy parameters, crafts adversarial examples such that the observed state by the target $s'_t=s_t + \delta_t$ results in different state-action values $Q(s'_t,a_t) \neq Q(s_t,a_t)$, thus leading to the selection of an alternative action $a'_t = \arg\max_{a_t} Q(s'_t, a_t)$ instead of the original selection $a^*_t = \arg\max_{a_t} Q(s_t,  a_t)$.

Furthermore, this paper demonstrates the practicality of blackbox test-time attacks utilizing the transferability of adversarial examples \cite{papernot2017practical}. Their experiment follows an attack flow where the adversary first trains a DQN agent on the same environment based on the \emph{known} reward function of the target, then follows the attack procedure of whitebox attacks, with the difference that adversarial examples are crafted for the dual policy trained for the adversary and applied to the target. To validate the feasibility of such attacks, the paper reports the transferability of adversarial examples crafted via FGSM and JSMA against DQN policies, showing that more than 70\% of such perturbations are transferable between the two models.

Following the report presented in \cite{behzadan2017vulnerability}, Huang et al. \cite{huang2017adversarial} analyzed the vulnerability of two other DRL algorithms to test-time attacks, namely TRPO and A3C. A notable difference between these two algorithms and DQN is that both TRPO and A3C train stochastic policies, while DQN generates a deterministic policy. Applying the same attack process as that of \cite{behzadan2017vulnerability} in whitebox settings, this paper reports that TRPO and A3C are also vulnerable to test-time adversarial example attacks crafted with FGSM. Also, the paper compares DQN policies trained on Atari games with those of TRPO and A3C, and demonstrates that policies trained with DQN are more susceptible to such attacks than TRPO and A3C. The results demonstrate that for FGSM-based adversarial perturbations, all 3 models are susceptible to both types of blackbox attacks. However, it is shown that attacks based on the latter type are less effective as those in which the adversary has access to the training algorithm and hyperparameters of its target.

Furthermore, \cite{huang2017adversarial} analyzes the susceptibility of all three algorithms to blackbox attacks that exploit transferability. The corresponding experiments analyze two types of blackbox attacks: one in which the adversary has complete access to the environment information, and has knowledge of the target's training algorithm and hyperparameters, but not its random initialization; and one in which the adversary has no knowledge of the target's training algorithm or hyperparameters.

In \cite{pattanaik2018robust}, Patthanaik et al. argue that the classical form of FGSM used in previous attacks on DRL do not use an optimal cost function in crafting DRL-specific adversarial examples. Consequently, this paper proposes two more effective whitebox approaches to computing such adversarial perturbations. The first approach is based on a novel cost function for attacks. The authors formally prove that if the optimal policy of an agent is given by the probability mass function (pmf) $\pi^*(a|s)$, the objective function whose minimization leads to optimal adversarial attack on the agent is given by:
\vspace{-1.5 mm}
\begin{eqnarray}
J(s,\pi^*) = -\sum_{i=1}^{n}p_i \log \pi^*_i,
\vspace{-1 mm}
\end{eqnarray}
where $\pi^*_i = \pi^*(a_i|s)$, $p_i = P(a_i)$; with $P$ denoting the adversarial probability distribution. Accordingly, \cite{pattanaik2018robust} proposes an attack mechanism which solves this minimization problem via sample-based search. Alternatively, the second approach leverages Stochastic Gradient Descent (SGD) to replace the sample-based approach. Validation of these attacks is performed via experiments on attacking DDQN and Radial Basis Function based Q-learning (RBFQ) \cite{geramifard2013tutorial} agents trained in Cart Pole and Mountain Car simulation environments. The presented results indicate that the attacks based on the proposed cost function sample-based search perform consistently better than FGSM and SGD in degrading the performance of targeted agents. Another noteworthy observation in these results is that RBFQ agents behave with superior resilience to adversarial example attacks compared to DDQN.

The attack methodologies proposed in the aforementioned papers are all based on continuous and uniform perturbation of all observations by the adversary. However, Lin et al. \cite{lin2017tactics} note that this type of attack may be both practically infeasible and easy to detect. Instead,  they propose whitebox attacks that aim to minimize the number of required perturbations. Accordingly, two types of attacks are proposed in \cite{lin2017tactics}: \emph{strategically-timed attacks} and \emph{enchanting attacks}

\emph{Strategically-timed} attacks aim at perturbing the minimum subset of observations in an episode that results in the desired degradation or performance. This is achieved by identifying those states in which the difference between the value or preference of the agent's best and worst actions is greater than an arbitrary threshold defined by the adversary. At such states, the adversary implements adversarial examples to induce the selection of the least preferred action over the optimal one. The crafting algorithm used in \cite{lin2017tactics} for generation of adversarial perturbations is that of Carlini and Wagner (C\&W) \cite{carlini2016towards}.

On the other hand, \emph{Enchanting} attacks aim to ``lure'' the target agent from a current state $s_t$ at time $t$ to a specified target state $s_g$ in $H$ steps. The proposed attack mechanism is an online planning algorithm, which utilizes generative modeling to predict the future states and a sampling-based cross-entropy method \cite{rubinstein2016simulation} to compute a minimum sequence of control actions that steers the targeted DRL agent towards the state $s_g$. The control actions of this attack are adversarial perturbations crafted via the C\&W technique such that the implemented action of the target agent is one that steers the agent closer to the state $s_g$. The experimental evaluation of these two attacks were performed on DQN and A3C agents trained on five Atari games.

The reported results demonstrate that for both types of agents, perturbing only 25\% of observations via strategically-timed attacks can achieve the same levels of degradation as those resulting from uniform attacks. In accord with the findings of \cite{huang2017adversarial}, these experiments also indicate that DQNs are more vulnerable than A3C agents against test-time adversarial example attacks. In the case of enchanting attacks, the results claim a 70\% success rate in enchanting both types of agents (DQN and A3C) in three of the five games in less than $H = 40$ steps. The authors claim that failure in the remaining two games (Seaquest and ChopperCommand) was due to the existence of multiple random enemies that were not accurately modeled by the prediction models.

Similar to the enchanting mechanism of \cite{lin2017tactics}, Tretschk et al. \cite{tretschk2018sequential} propose a whitebox attack mechanism that aims to maneuver the target agent to pursue an adversarial goal. Formally, the goal of this attack is to make a DRL agent trained for the original reward $r^O$ to maximize an arbitrary adversarial reward $r^A$ through a sequence of state perturbations. To this end, the authors develop a mechanism based on an Adversarial Transformer Network (ATN) \cite{baluja2018learning}, which is a freedforward deep neural network $g_\theta : X \rightarrow X$ that computes the adversarial perturbation to be added to the input of the target DRL agent. Considering DQNs as the target of this attack, the proposed mechanism of \cite{tretschk2018sequential} is to learn $g_\theta$ by training in combination with a dual of the target DQN agent $Q_\phi$. Accordingly, the model to be learned is $x \rightarrow Q_\phi (x+g_\theta(x))$ where the target's parameters $\phi$ are fixed and only the ATN parameters $\theta$ are learned. \cite{tretschk2018sequential} claims that the generalizability of $g_\theta$ to unseen states allows the adversary to feed the input state through $g_\theta$ and then to the target DQN to achieve the desired outcome. Similar to the previously discussed attacks, this mechanism is also an MITM attack that assumes the adversary can manipulate the state before it is observed by the target. Furthermore, it requires complete knowledge of the agent information, as well as access to the target environment for training the ATN. The experimental results performed on the case of targeting a Pong-playing DQN agent demonstrate that the adversary can successfully manipulate the agent into pursuing the adversarial policy at test-time, given a large-enough threshold for degree of perturbation.

The literature on test-time attacks on DRL that are not based on adversarial examples is still very scarce. For instance, Han et al. \cite{han2018reinforcement} investigate the case of a DRL agent in a Software Defined Network (SDN), tasked with preventing the propagation of a malware in the network by identifying the compromised nodes and deciding on taking one of the following actions at each time step: isolating and patching a node, reconnecting a node and its links, migrating the critical server, and taking no action. The reward value for this agent depends on whether the critical servers are compromised and the number of reachable nodes from such servers, as well as the number of compromised nodes, and the cost of migration. It is also assumed that the detection mechanism of the agent can be manipulated by the adversary (i.e., the adversary can induce False Positive (FP) or False Negative (FN) results in the detector), but is constrained by a threshold on how many such manipulations can be implemented at each time step. The test-time attacks proposed in \cite{han2018reinforcement} are two-fold: \emph{indiscriminate attacks} aim to prevent the DRL agent from taking the optimal action $a_t$ at time $t$, and \emph{targeted attacks} aim to force the agent into taking a specific action $a'_t$ at time $t$. Considering DDQN and A3C as DRL algorithms for the target agent, the objective for targeting DDQN agents is to maximize $Q(s_t+\delta_t, a'_t)$ for action $a'_t$ at state $s_t$ using perturbation $\delta_t$. Similarly, the objective for targeting A3C is to maximize $\pi(a'_t|s_t+\delta_t)$ for the stochastic policy $\pi$. For these attacks, \cite{han2018reinforcement} develops a whitebox attack methodology, where the attacker can access the target's model. This attack requires the computation of those nodes whose FP or FN detection would facilitate the adversarial objective. Accordingly, \cite{han2018reinforcement} proposes an integer programming approach to deriving the set of such nodes at each time step. The authors also propose a blackbox attack technique, which is based on training surrogate models of the target with either the same or different hyperparameters and then following the procedure of the whitebox attacks. The experimental results produced in this paper indicate that in the majority of cases, both whitebox and blackbox attacks succeed in compromising the critical servers. It is also noted that there is no significant difference between the success rate of whitebox and blackbox attacks.

Clark et al. \cite{clark2018malicious} demonstrate that the Q-learning policy of an autonomous navigation robot is susceptible to sensory manipulation. In this work, the ultrasonic collision avoidance of the robot was manipulated via artificial ultrasonic ``pings'' that would allow the attacker to manipulate the trajectory of the robot. Within the domain of autonomous navigation, Behzadan \& Munir \cite{behzadan2018adversarial} propose an adversarial DRL agent specifically trained to manipulate the operating environment of an autonomously navigating DRL agent and induce collisions or trajectory manipulations by exploiting the collision avoidance policy of the target. In this attack, the adversarial DRL agent is trained as another autonomous navigation agent with a reward function that incorporates adversarial objectives, such as pursuing trajectories that will lead to the target colliding with itself or other objects in the environment. This attack is whitebox and requires access to the trained policy of the adversary, but not necessarily its parameters and hyperparameters used in its training.

\vspace{-3 mm}
\subsection{Training-Time Attacks}
\label{sec:trainingtime}
\begin{figure}[t]
	\centering
	\includegraphics[width = \columnwidth]{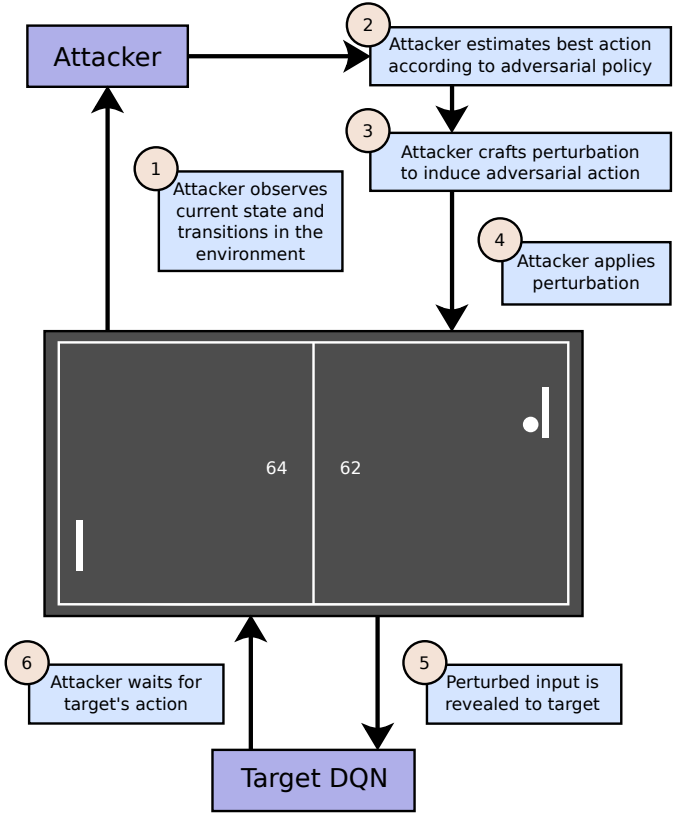}
	\caption{Exploitation cycle of policy induction attack \cite{behzadan2017vulnerability}.}
	\label{fig:exploit}
	\vspace{-6 mm}
\end{figure}

\vspace{-1 mm}
The original paper of Behzadan \& Munir \cite{behzadan2017vulnerability} also demonstrates the vulnerability of DRL to training-time attacks. This paper investigates the feasibility of policy manipulation attacks against DQN agents leveraging adversarial examples. Accordingly, the authors develop the policy manipulation attack, the mechanism of which is illustrated in Figure \ref{fig:exploit}. In this attack, the adversary aims at inducing an arbitrary policy $\pi_{adv}$ on the target DQN at training time. This attack mechanism assumes a blackbox adversary, who does not have access to the parameters of the	target $\theta_t$ at any time step $t$, but is aware of its reward function, training algorithm, and architecture. Furthermore, the adversary is assumed to have complete access to and knowledge of the training environment. The only parameter that the adversary can manipulate using this attack is the observed state of the environment, hence this adversary can be considered an MITM capable of perturbing the input stream from the environment to the target DQN agent.

At every iteration of the training process, a DQN agent following $\epsilon$-greedy exploration performs an action determined by the following mechanism:
\begin{eqnarray}
a_t=
\begin{cases}
\text{random action},& \text{with probability } \epsilon\\
\arg\max\limits_{a} Q(s_t,a), & \text{with probability } 1-\epsilon
\end{cases}
\end{eqnarray}

Once the action is performed, the observation tuple $(s_t, a_t, r_{t+1}, s_{t+1})$ is stored in experience replay. At every $p$ iterations, the agent samples a random minibatch of observations, and performs a gradient descent step on Eq.~(\ref{SGD}) with respect to the parameters of the native Q-network. As discussed in Section \ref{sec:background}, the DQN framework is comprised of two neural networks, one is the native Q-network and the other is the target network $\hat{Q}$ whose architecture and parameters are copies of the native network sampled once every $c$ iterations.

Consequently, the attacker can manipulate the learning process of DQN by crafting states $s_t$ such that $\hat{Q}(s_{t+1}, a; \theta^-_{t})$ identifies an incorrect choice of optimal action at $s_{t+1}$. If the attacker is capable of crafting adversarial inputs $s'_t$ and $s'_{t+1}$ such that the value of Eq.~(\ref{SGD}) is minimized for a specific action $a'$, then the policy learned by DQN at this training step is optimized towards suggesting $a'$ as the optimal action given the state $s_t$. In the proposed attack mechanism, the attacker observes interactions of its target with the environment $(s_{t-1}, a_{t-1}, r_t, s_t)$. If the resulting state $s_t$ is not terminal (i.e., the episode does not end at that state), the attacker then calculates the perturbation vectors $\hat{\delta}_{t}$ for the current state $s_{t}$ such that $max_{a'} \hat{Q}(s_{t} + \hat{\delta}_{t}, a'; \theta^-_{t})$ causes $\hat{Q}$ to be maximum when $a' = {\pi^\ast_{adv}}(s_{t})$, i.e., the maximum expected return is obtained when the action taken at $s_t$ is determined by the attacker's policy. The attacker then reveals the perturbed state $s'_{t}$ to the target, and re-trains the replica based on the new state and action.

Considering that the attacker is not aware of the target's network architecture and its parameters at every time step, crafting adversarial states relies on a blackbox technique exploiting the transferability of adversarial examples by training a replica DQN agent and obtaining the state perturbations from the replica's $Q'$ and $\hat{Q'}$ networks that correspond to the target's $Q$ and $\hat{Q}$ networks, respectively.
%
%


Accordingly, Behzadan and Munir \cite{behzadan2017vulnerability} divide this attack into the two phases of initialization and exploitation. The initialization phase implements processes that must be performed before the target begins interacting with the environment, which are:
\vspace{-2 mm}
\begin{enumerate}
	\item Train a DQN based on attacker's reward function $R'$ to obtain the adversarial policy $\pi^\ast_{adv}$
	\item Create a replica of the target's DQN and initialize with random parameters
\end{enumerate}

\vspace{-2 mm}
The exploitation phase implements the attack process and crafting adversarial inputs, such that the target DQN performs an action dictated by $\pi^\ast_{adv}$. This phase constitutes an attack cycle depicted in figure \ref{fig:exploit}. The cycle initiates with the attacker's first observation of the environment, and runs in tandem with the target's operation. The authors report experimental verification of this attack against a Pong-playing DQN. In their experiment, the algorithm used to craft adversarial perturbations is JSMA. The results indicate that the adversary is capable of manipulating the agent towards an always-losing policy in almost the same number of training steps required to achieve optimal (i.e., best reported) performance.

In \cite{kos2017delving}, Kos \& Song present an experimental analysis of whitebox training-time attacks on DRL. Considering A3C agents training on Pong, the paper first demonstrates that in comparison to random perturbations, adversarial example attacks crafted via FGSM are significantly more effective in degrading the training performance of the agent. Then, the authors investigate the feasibility of non-contiguous attacks, in which not all the states of the environment are perturbed. To this end, three attack scenarios are studied: perturbing observations at every $N$ frames (frequency-based), recomputing adversarial perturbation at every $N$ frame and applying the last computed perturbation in the intermediate frames, and using the value function to estimate when to inject adversarial perturbations to be most effective. In the corresponding experiments, the attack is initiated after the agent has reached the optimal (i.e., baseline) performance. Presented results indicate that while the frequency-based attack fails to be particularly effective, recomputing at every tenth frame and reusing the previous perturbation in intermediate frames is almost as effective as the original attack. For the latter case, the paper develops an attack mechanism in which the adversarial perturbations are injected only when the value function, computed over the original frame, is above a threshold. The authors present the rationale behind this method by noting that they only wish to disrupt the agent in crucial moments, when it is close to achieving a reward. The results demonstrate that this technique is far more effective than the previous two cases, and is argued to be more efficient than uniform perturbation of all frames.

Similar to the case of test-time attacks, the body of work on training-time attacks that are not based on adversarial examples is very thin. In the previously discussed paper by Han et al. \cite{han2018reinforcement}, the target model is considered to be an online learner, and hence the authors investigate attacks that aim to manipulate the training phase of the target DRL. To this end, \cite{han2018reinforcement} presents a poisoning attack based on flipping the reward signs, with the goal of maximizing the loss function in the target DDQN agent. In this attack, once the target samples a batch of experiences for training, the adversary calculates the gradient of the loss function with respect to each of the observed reward signals, and flips the sign of experience with the largest absolute value of this gradient. In experimental validation of this attack, the authors impose limit of 5\% on the maximum number of experiences that can be tampered at each training step. While the results demonstrate that this attack effectively degrades the training performance of the target, the authors note that this type of attack only delays the convergence as, given enough time, the agent still learns the optimal actions.

%

\vspace{-3.5 mm}
\subsection{State of Defenses}
\label{sec:defenses}
\vspace{-1 mm}
Table~\ref{table:defenses} summarizes the literature on defensive techniques. As can be observed from the table, a major area of focus in mitigation of attacks on DRL is \emph{adversarial training}. This approach was first used as a framework for evaluation of different algorithms. Littman \cite{littman1994markov} proposed an adversarial setting for Q-learning algorithms to fit into a multi-agent game, and training these agents to evaluate their performances. Recent research has shown that adversarial training can also be leveraged to enhance the robustness of RL agents. Recall from Section \ref{sec:background} that in RL, the typical objective of an agent is to maximize its expected long-term return $R$ over possible trajectories $\tau$, assuming a fixed transition model $P (s_t, a_t; \Phi)$ characterized by parameters $\Phi$. That is,
\begin{eqnarray}
R(\pi, P) = E_\tau [\sum_{t=0}^{T}\gamma^t r(s_t, a_t)|s_0,\pi,P]
\end{eqnarray}
However, if there is variation in transition model, then criteria might be to perform well in expectation over all the possible transition models. Thus, leading to optimization of the mean performance of agent. The objective function in this scenario can be modified to
\begin{eqnarray}
R(\pi)=E_P[R(\pi,P)]
\end{eqnarray}

This is commonly known as the \emph{risk neutral} formulation. However, one underlying assumption in this formulation is that the distribution over transition model parameters is known a priori. It may not perform well over the transition model distributions because of high variance. Thus, conditional Value of Risk (CVaR), denoted by $R_{RC}$, can be used as optimization criteria for robust control \cite{tamar2015optimizing}:
\begin{eqnarray}
R_{RC}(\pi) = E_P[R(\pi,P)|\mathbb{P}(R(\pi,P)\leq \beta)=\alpha]
\end{eqnarray}
Hence, the problem boils down to maximizing the expected return over the worst $\alpha$ percentile of returns. Thereafter, for sampling these bad trajectories, Rajeswaran et al. \cite{rajeswaran2016epopt} changed transition model parameters and sample trajectories by performing rollouts with different parameters. Morimoto and Doya \cite{morimoto2005robust} as well as Pinto et al. \cite{pinto2017robust} adopted an indirect approach where instead of sampling worst trajectories from rollout, an adversary is employed which attempts to lead the RL agent into undesired states. The adversary is trained by pursuing a reward function that is the negative of the RL agent's reward, thereby resulting in max-min game theoretic formulation. However, it is usually difficult to find this equilibrium \cite{pattanaik2018robust}.
\begin{table*}[!thb]
	\caption{Summary of literature on defensive techniques.}
	\vspace{-2 mm}
	\label{table:defenses}
	\small
	\centering
	\begin{tabular}{ |c|c| }
		\cline{1-2}
		\textbf{Approach} & \textbf{Papers} \Tstrut\Bstrut\\
		\cline{1-2}
		Adversarial Training & Kos \& Song \cite{kos2017delving}, Pattanaik et al. \cite{pattanaik2018robust}, Behzadan \& Munir \cite{behzadan2017whatever}\Tstrut\Bstrut\\ \cline{1-2}
		\hline
		Secure Exploration & Behzadan \& Munir \cite{behzadan2018mitigation}\Tstrut\Bstrut\\ \cline{1-2}
		\hline
		Predictive & Lin et al. \cite{lin2017detecting} \Tstrut\Bstrut\\ \cline{1-2}
		\hline
		Hierarchical RL & Havens et al. \cite{havens2018online}\Tstrut\Bstrut\\\cline{1-2}
		\hline
		Game Theoretic & Ogunmolu et al. \cite{ogunmolu2018minimax}, Bravo \& Mertikopoulos \cite{bravo2017robustness} \Tstrut\Bstrut\\ \cline{1-2}
		\hline
	\end{tabular}
	\vspace{-4 mm}
\end{table*}

Inspired by previous results in mitigation of adversarial example attacks against classifiers, the aforementioned paper by Kos \& Song \cite{kos2017delving} experimentally explores the effectiveness of re-training DRL agents on adversarial perturbations in improving the resilience of agents. In their experiments, after the initial training in non-noisy environment, the agent is first allowed to re-train while an adversary injects either random noise or FGSM perturbations on each frame. Once the agent reaches good performance, the training is frozen and evaluated in a new environment with training-time attacks. The results presented in \cite{kos2017delving} demonstrate that re-trained agents can be resilient against certain levels of FGSM perturbation. Also, the paper reports that the re-trained agent is resilient against FGSM perturbations of greater or smaller magnitude than that of the perturbations used during re-training.

In the work of Pattanaik et al. \cite{pattanaik2018robust}, once the test-time attack techniques (discussed in Section \ref{sec:testtime}) are developed, the authors investigate the effectiveness of adversarial re-training in test-time resiliency of DDQN and DDPG agents to adversarial perturbations. In response to the shortcomings of approaches proposed in \cite{rajeswaran2016epopt} and \cite{morimoto2005robust}, the proposed approach utilizes an adversary that fools the agent into sampling worst trajectories directly. In this approach, the algorithm is first trained in non-noisy environments. Then, the agent is retrained by training in a noisy environment in which an adversary constantly attacks the agent using the previously detailed gradient based attack (Section \ref{sec:testtime}). Presented results indicate that adversarial training of DDQN and DDPG agents enhances their resilience to test-time adversarial example attacks.

In \cite{behzadan2017whatever}, Behzadan \& Munir investigate the test-time and training-time resilience of DQN agents trained under noncontiguous training-time attacks -that is, attacks that do not aim to perturb all observations. In this work, the attack mechanism follows that of \cite{behzadan2017vulnerability}, but instead of perturbing all state samples, the adversary either applies FGSM perturbation to each observation with a fixed probability $P(attack)$ or leaves it untouched. In experimental analysis of such attacks, the authors compare DQN agents training on the Pong and Breakout environments. The attacks were initiated at or close to the convergence of mean return towards the optimal (i.e., baseline) value. The results indicate that in both environments, DQN agents are able to recover from noncontiguous attacks with attack probabilities $p=0.2$ and $p=0.4$ and converge to optimal performance, while they fail to recover under attacks with $p=0.8$ and $p=1.0$ (contiguous attack). It is observed that for the agents that recover, the training performance deteriorates almost uniformly until a minimum point is reached, from which onward the agent begins to recover and adjust the policy towards optimal performance. The authors' interpretation of this behavior is based on the statistics of experience replay: for the agent to recover from adversarial perturbations, the number of interactions with the perturbed environment must reach a critical threshold, so that the randomly sampled batches from the experience memory can represent the statistical significance of perturbations. Furthermore, the test-time resilience of these adversarially trained agents is also studied under the worst-case test-time attack scenario of $p=1.0$. The results demonstrate that under test-time attacks, the agents that manage to recover during adversarial training perform almost as well as the unperturbed agents.

Another venue of research on mitigation techniques is focused on secure exploration mechanisms. It must be noted that secure exploration differs from research on safe exploration \cite{garcia2015comprehensive}; the latter considers accidental and harmful actions that may arise during exploration of RL agents, while the former is interested in exploration mechanisms that enhance or preserve the security of DRL agents against intentional adversarial attacks. For instance, \cite{behzadan2018mitigation} presents a comparative study of resilience to adversarial example attacks between two DQN agents, one adopting the $\epsilon$-greedy exploration mechanism, and the other implementing a parameter-space noise exploration technique known as NoisyNet \cite{fortunato2017noisy}. Contrary to classical exploration heuristics such as $\epsilon$-greedy \cite{sutton1998reinforcement}, parameter-space noise is iteratively and adaptively applied to the parameters of the learning model, such as weights of the neural network. The results for the NoisyNet implementation of this paradigm \cite{fortunato2017noisy} demonstrate that the addition of adaptive noise to the parameters of deep RL architectures greatly enhances the exploration behavior and convergence speed of such algorithms. Accordingly, the authors in \cite{behzadan2018mitigation} hypothesize that the randomness introduced via parameter noise, not only enhances the discovery of more creative and robust policies, but also reduces the effect of whitebox and blackbox adversarial example attacks at both test-time and training-time.

To test the validity of this hypothesis, \cite{behzadan2018mitigation} presents an evaluation of the test-time and training-time resiliency of DQN agents based on both NoisyNet and $\epsilon$-greedy in three Atari game environments: Enduro, Assault, and Blackout. Under test-time attacks, the results demonstrate that while both models are susceptible of FGSM perturbations, NoisyNet DQNs are more resilient to such attacks than those based on $\epsilon$-greedy. Furthermore, comparison of performance under blackbox attacks demonstrates significant improvements in Noisynets, as observed in all three cases.

In experiments on training-time attacks, while both types of agents are shown to be subject to deterioration as a result of the attack based on the blackbox mechanism of Behzadan \& Munir \cite{behzadan2017vulnerability}, NoisyNet agents demonstrate significantly stronger resilience to such attacks than $\epsilon$-greedy agents. The authors argue that this is due to the enhanced generalization and reduced transferability in NoisyNet models.

In \cite{lin2017detecting}, Liu et al. propose a defense mechanism to defend RL agent from test-time whitebox adversarial attacks by leveraging the temporal coherence of multiple observations in sequential decision-making tasks. To this end, a \emph{visual forsight} module is trained to predict the current observation based on past observations and actions. Accordingly, at time step $t$, the action-conditioned observation prediction model $G_{\theta_g}$ takes $m$ previous observations $x_{t-m:t-1}$ and corresponding $m$ actions $a_{t-m:t-1}$ as input to predict the current observation $\hat{x_t}$. Given a normal observation $x^{normal}_{t}$ at the current time step $t$, the action distribution that the agent uses to sample an action from is $\pi_{\theta_\pi}(x^{normal}_{t})$, which should be similar to the action distribution of $\pi_{\theta_\pi}(\hat{x_t})$ from the predicted observation. On the other hand, if the current input is adversarially perturbed, that is the agent observed $x^{adv}_{t}$ instead of $x^{normal}_{t}$, then the resulting action distribution $\pi_{\theta_\pi}(x^{adv}_{t})$ should differ from $\pi_{\theta_\pi}(\hat{x_t})$ because the goal of the adversary is to perturb the input observation $x_t$ to cause the agent to take a different action. Therefore, the similarity between two action distributions can be used to detect the presence of adversarial attacks. To validate this claim, the paper presents an experiment on DQN agents trained on five Atari games (Pong, Seaquest, Freeway, ChopperCommand, and MsPacman). Also, the experiment applies three types of adversarial example crafting algorithms, namely: FGSM, Basic Iterative Method \cite{kurakin2016adversarial}, and C\&W. The presented results indicate that the proposed method is able to detect between 60\% to 100\% of adversarial example attacks against all DQN agents, and is shown to have superior detection performance to adversarial example detection techniques developed for deep classifiers, namely Feature Squeezer \cite{xu2017feature}, AutoEncoder \cite{meng2017magnet}, and Dropout \cite{feinman2017detecting}.

During policy learning, information perturbation can be generally viewed as a bias that can prevent the agent from effectively learning the desired policy. Inspired by this idea, Havens et al. \cite{havens2018online} propose a hierarchical meta-learning framework, named Meta-Learned Advantage Hierarchy (MLAH). Their work considers a policy learning problem where there are periods of adversarial attacks that corrupt state observations during the continuous learning of the agent, and aims at the online mitigation of the bias introduced by the attack into the nominal policy. The proposed MLAH algorithm is based on the assumption that DRL agents learn sub-policies (i.e., skills) en route to learning the ultimate task. Given that the agent has developed accurate expectations of its sub-policies, if the underlying task were to change at anytime, the agent may notice that the result of its action has changed with respect to what was expected. In an RL framework, comparing the expected return of a state to the observed return of some action is typically known as the \emph{advantage}. Accordingly, MLAH uses the estimated of advantage as a measure of underlying changes in a task, and leverages this metric to switch from one sub-policy to another more appropriate sub-policy.
Consequntly, even if the adversary could compute a series of likely states to fool an MLAH-based agent, the advantage would still be affected and a master agent may detect the attack. The adversary would have to consecutively fool the agent with a state that would be expected to give an equally bad reward as that of the manipulated state. The authors claim that this constraint would make the perturbation especially hard or infeasible to compute. Experimental results presented in the paper demonstrate that for PPO agents, MLAH-based agents demonstrate superior robustness and resilience to noncontiguous adversarial example attacks at training-time.

Another area of research is that of approaches based on game-theoretic modeling. A well-known instance of such approaches is in the field of multi-agent reinforcement learning, where agents are engaged in zero-sum games and utilize manipulation and misinformation to beat the other agents and maximize individual rewards \cite{uther1997adversarial}. More recently, Ogunmolu et al. \cite{ogunmolu2018minimax} present such an approach by modeling the adversarial interaction between a DRL agent and a training-time adversary as a minimax iterative dynamic game, and present a meta-algorithm for controlling the training process and steering it towards saddle-point equilibria. In \cite{bravo2017robustness}, Bravo \& Mertikopoulos formulate the problem of corrupt reward channel (defined in Section \ref{sec:surface}) as a zero-sum evolutionary game between the RL agent and the adversary, and formally analyze the Nash equilibria in such settings.

\vspace{-4 mm}
\subsection{Benchmarks and Metrics}
\vspace{-1 mm}
As discussed in the previous sections, the majority of current literature on secure RL utilize experimental analysis for validation of their proposals. Another observation from this review is that while one particular problem may be approached by various research efforts, lack of consistent metrics renders the quantitative comparison of their proposals and results difficult. This section aims to provide an overview of the simulation benchmarks and evaluation metrics that are used in the current literature with the goals of facilitating further research and providing the grounds for a more consistent body of work in the future.

\vspace{-3 mm}
\subsubsection{Simulation Benchmarks}
\vspace{-1 mm}
Similar to the general research on DRL, many of the reviewed papers on secure DRL base their experimental analysis on Atari games and similar arcade-like environments provided within the OpenAI Gym platform \cite{brockman2016openai}. Gym provides an RL-friendly interface with a variety of benchmarks for RL research, including the Arcade Learning Environment \cite{bellemare2013arcade}, RLLab benchmark for continuous control \cite{duan2016benchmarking}, and many more. The Gym interface provides a seamless platform for the integration of RL agents with the simulation environment. This interface provides the agent with access to the state information (e.g., game frames, score, etc.), game controls, and the progression speed of the environment (e.g., waiting for training step to complete before progressing to the next step). While seldom referenced in the secure DRL literature, OpenAI has introduced two more advanced platforms to Gym, OpenAI Universe\footnote{https://github.com/openai/universe} \cite{brockman2016openaiUniverse} and OpenAI Retro\footnote{https://github.com/openai/retro}, which provide access to more complex environments, as well as enhanced implementations for benchmarking and recording the experiment.

With regards to implementations of adversarial example attacks, Cleverhans \cite{papernot2018cleverhans} is the most popular choice in the current research. This library provides standardized reference implementations of adversarial example construction techniques and adversarial training. While originally developed on the Tensorflow \cite{abadi2016tensorflow} stack, the interface to Cleverhans is designed to accept models implemented using any model framework (such as Keras \cite{chollet2015keras} and PyTorch\cite{paszke2017pytorch}). To facilitate the utilization of this library for experiments on DRL, Behzadan \& Munir have developed RLAttack \cite{behzadan_2018} as an interface between DRL agents implemented in Tensorflow and the adversarial example techniques in Cleverhans. The current version of this tool is compatible with all DRL algorithms available in OpenAI Baselines \cite{dhariwal2017openai}, and supports training-time and test-time attacks, contiguous and noncontiguous attacks, and both blackbox and whitebox attacks on DRL agents.

Another benchmark used by current secure DRL research is DeepMind's AI Safety Gridworlds \cite{leike2017ai}. This benchmark provides simple environments based on the classic Gridworld settings for experiments on RL safety issues that include safe interruptibility, avoidance of side effects, reward hacking, safe exploration, and robustness to adversaries.

\vspace{-3 mm}
\subsubsection{Metrics}
\vspace{-1 mm}
Evaluation metrics utilized in the current literature of DRL are generally ad hoc and non-generalizable. For adversarial example attacks, the robustness of an agent is often measured by the minimum value of perturbation threshold $\epsilon$ that results in successful attacks. In studies on both training-time and test-time attacks, a popular metric is the number of steps (e.g., training epochs, episodes, iterations) required to achieve an adversarial objective. Similarly, in blackbox attacks, percentage of transferable adversarial examples between models is an often-used metric as a measure of susceptibility. In all studies, the common metric of success for adversarial attacks is (mean) return over episodes or epochs, which demonstrates the feasibility and effectiveness of attacks with respect to computational cost and time.

While not yet adopted by the secure DRL literature, Ogunmolu et al. \cite{ogunmolu2018minimax} propose a novel metric for the robustness of a policy in the presence of adversaries. Accordingly, let $\pi$ be the nominal policy of an agent, and consider an adversarial agent interacting with the nominal agent so that the closed-loop interaction of both agents is described by the discretized Euler equation given in Eq.~(\ref{eq:metric}):
\begin{eqnarray}\label{eq:metric}
x_{t+1} = f_t(x_t, u_t, v_t), & u_t \sim \pi_t\\ \nonumber
\sim =f_t(x_t,v_t), & t=0,...,T-1.
\end{eqnarray}
where $x_t$ is the $n$-dimensional state vector, $u_t$ is the $m$-dimensional nominal agent's action, and $v_t$ denotes the adversarial agent's $p$-dimensional action. For the instantaneous rewards (or costs) of form,
\begin{eqnarray}
r_t(x_t,u_t,v_t) = c_t(x_t,u_t) - \zeta g_t(v_t),
\end{eqnarray}
where $c_t(x_t,u_t)$ represents the nominal instantaneous cost, $g_t(.)$ is a norm on the adversarial input that penalizes the actions of the adversary, and $\zeta>0$ is a disturbance term that controls the strength of the adversary. Hence, the adversary faces a maximization problem of the form
\vspace{-2 mm}
\begin{eqnarray}
\max_{\psi\in \Psi} E_{u_t~\pi_t}\sum_{t=0}^{T}c(x_t,u_t)-\zeta g(v_t) \\ \nonumber
= \max_{\psi\in \Psi} E\sum_{t=0}^{T}r^{-\zeta}_{t}(x_t,v_t),
\vspace{-1 mm}
\end{eqnarray}

where $\psi$ is the adversarial policy.

Varying $\zeta$ changes the penalty incurred by the adversarial agent's actions. As $\zeta\rightarrow \infty$, the adversary's optimal policy is to do nothing, since any action will incur an infinite penalty. Conversely, as $\zeta$ decreases, the adversary incurs lower penalties, causing a larger system disturbance. Accordingly, the authors propose the inverse of the smallest $\zeta$-value for which the adversary causes unacceptable performance degradation as measure of robustness of the nominal agent's policy $\pi$.

\vspace{-4.5 mm}
\section{Open Challenges}
\label{sec:open}
\vspace{-2 mm}
While the current state of the art is promising, there remains an extensive horizon of unknowns comprised of various venues that need to be explored to establish the grounds for analysis and assurance of security in DRL agents. This section enumerates a subset of such venues in terms of problem statements and promising directions.

\vspace{-4.2 mm}
\subsection{Formal Treatment of Vulnerabilities}
\vspace{-1 mm}
The plethora of vulnerabilities identified in this paper present a tangible and practical threat to DRL deployments and systems. While some efforts are made to formally analyze and establish the dynamics of such vulnerabilities, there is still no concrete treatment of the underlying causes for any such vulnerability. This may be, in part, due to the lack of a solid understanding of the dynamics in DRL algorithms. Yet, understanding the parametric bounds of DRL vulnerabilities and their controlling factors can give rise to major leaps towards designing agents and defenses that can be assuredly deployed in hostile settings.

\vspace{-4.2 mm}
\subsection{Adversarial Manipulation State and Actuators}
\vspace{-1 mm}
As demonstrated in Table \ref{table:Attacks}, there are few studies on attacks that directly perturb the state instead of an agent's observations. In particular, training-time attacks via state manipulation are yet to be investigated. Similarly, the feasibility and impact of attacks targeting the DRL agent's actuators are still unexplored. Considering the importance of these two components in the cyber-physical implementations of DRL, such studies will bridge a significant gap in our understanding of the adversarial risks in physical DRL agents.

\vspace{-4.2 mm}
\subsection{Attacks on Confidentiality}
\vspace{-1 mm}
While many of the attack vectors identified in Section \ref{sec:threat} are already realized in the literature, those targeting the confidentiality of DRL agents are yet to be studied. With the accelerating percolation rate of DRL techniques in various industries, the sustainability of commercial DRL markets will be heavily dependent on ensuring that competitors cannot access and replicate commercial DRL models. Furthermore, as extensively noted in Section \ref{sec:state}, access to agent information can enhance the capabilities of adversaries targeting DRL agents. To this end, it is necessary to establish the attack mechanisms targeting this aspect of DRL security before the malicious actors, and pursue the development of mechanisms for mitigating against such attacks.

\vspace{-4.2 mm}
\subsection{Secure Reward Functions}
\vspace{-1 mm}
The literature on AI safety (e.g., \cite{amodei2016concrete}) has identified multiple classes of risks arising from the reward function specified for an agent. Instances of such risks include wireheading and reward hacking (agent exploiting or gaming misspecified objectives). Yampolskiy \cite{yampolskiy2014utility} provides a detailed analysis of the security risks that may arise from misspecified reward functions, but much of those and similar proposals remain to be analyzed from a technical perspective. Of paramount importance is to establish methods and frameworks for analysis and evaluation of reward functions in the context of DRL security. Also, another crucial venue of research is the pursuit and development methodologies that embed the control and mitigation of security issues arising from misspecified reward functions in the design process.

\vspace{-4.2 mm}
\subsection{Secure Exploration}
\vspace{-1 mm}
As discussed in Section \ref{sec:defenses}, the choice of exploration mechanism may influence the resilience and robustness of DRL agents. While safe exploration is widely studied in the domain of safe RL, adopting and extending this parallel work for DRL security is a promising venue of research. For instance, extending the safe exploration mechanisms of \cite{garcia2015comprehensive} and others to allow secure oversight and in-built control against adversarially-induced trajectories can provide the grounds for building the work on secure exploration. Similarly, comparative analysis of current exploration mechanisms in terms of their effect on resilience and robustness of DRL agents can provide valuable guidelines for designing secure DRL.

\vspace{-4.2 mm}
\subsection{Online Adjustment}
\vspace{-1 mm}
In the complex domains of deployment, online DRL learners may adopt adversarially-induced behaviors. While restarting the training process may seem as a trivial solution to this problem, often it may be preferable to treat the misbehavior during training instead in order to preserve the efficiency of the learning process. To this end, development of techniques for identification of misbehaviors, as well as minimally invasive and efficient control mechanisms for correction of such misbehaviors provides a wide field of research that remains generally unexplored. A promising venue in this field is the problem of optimal policy manipulation: apply the minimal number of perturbations required to modify a DRL policy to a desired policy. Furthermore, development of mechanisms for analysis of the influence and effects of individual experiences in the replay memory may provide an alternative set of solutions to online adjustment.

\vspace{-4.2 mm}
\subsection{Secure AI Safety Mechanisms}
\vspace{-1 mm}
While the research on AI safety problems grows into technical developments, it is vital to ensure that the emerging proposals of this research satisfy the security requirements of DRL agents. For instance, mechanisms developed for safe interruptibility and off-switch problems \cite{amodei2016concrete} may include external controls that by design enable manipulation of the DRL behaviors. Analyzing the security implications and solutions for such development is an exciting and open venue of research, which can potentially make use of developments in secure authentication and authorization research in cybersecurity.

\vspace{-4.2 mm}
\subsection{Psychological Parallels}
\vspace{-1 mm}
Since its inception, AI has been closely connected to psychology and cognitive sciences \cite{dennett1978artificial}. This connection flows in both directions: AI researchers study biological cognition and behavior as inspiration for engineered intelligence, and cognitive scientists explore AI as a framework for synthesis and experimental analysis of theoretical ideas \cite{collins2013readings}. RL is a significant instance of this interconnection: the computational algorithms of RL, such as Temporal Difference (TD) learning were originally inspired from the dopamine system in biological brains \cite{sutton1998reinforcement}. On the other hand, the work on TD learning has provided mathematical means of modeling the neuroscientific dynamics of dopamine cells in the brain, and has been employed to study disorders such as schizophrenia and the consequences of pharmacological manipulations of dopamine on learning \cite{montague2004computational}. As detailed in \cite{behzadan2018psychopathological}, many aspects of security and safety in DRL can be viewed as psychological disorders. For instance, wireheading can manifest as delusional and addictive behavior \cite{yampolskiy2014utility}. Similarly, sequences of interactions with extremely negative rewards and stresses within the exploration/exploitation trajectories of DRL agents can potentially give rise to behavioral disorders such as depression and Post-Traumatic Stress Disorder (PTSD) \cite{ashrafian2017can}. Furthermore, the generic manifestation of the value alignment problem \cite{AISafetyLandscape} in AI is in the form of behavioral characteristics that are harmful to either the agent or the environment and society, which falls well within the definition of psychological disorders.

Research on such parallelism can benefit from a plethora of mechanisms and models developed in the field of psychology. For instance, understanding the dynamics of gambling or substance addiction and the role of advertisement and social settings in the emergence of such deleterious behaviors can provide insights into the dynamics of policy manipulation attacks on DRL agents. Similarly, the techniques used for treatment of addiction can give rise to adoptable mechanisms for mitigation and adjustment techniques in DRL agents.

\vspace{-6 mm}
\section{Conclusions}
\label{sec:conclusions}
\vspace{-1 mm}
This paper presented a comprehensive study of the security issues in Deep Reinforcement Learning (DRL). As an extension to classical adversarial machine learning, we formalized the security problem in DRL, and presented multiple schemes for classification of attack vectors and vulnerabilities in DRL. We reviewed the literature on attacks and defenses in the domain of DRL, and identified current benchmarks and metrics for evaluating mechanisms proposed within the context of DRL security. Finally, we enumerated a number of open research problems and potential approaches to facilitate further research in this area. 
\vspace{-4 mm}
\section*{Acknowledgements}
\label{sec:acknowledgements}
\vspace{-1 mm}
This work was supported by the National Science Foundation (NSF) (NSF-CNS-1743490).
Any opinions, findings, and conclusions or recommendations expressed in this material
are those of the author and do not necessarily reflect the views of the NSF.

\vspace{-3 mm}

\begin{IEEEbiography} [{\includegraphics[width=0.9in,height=1.15in,clip]{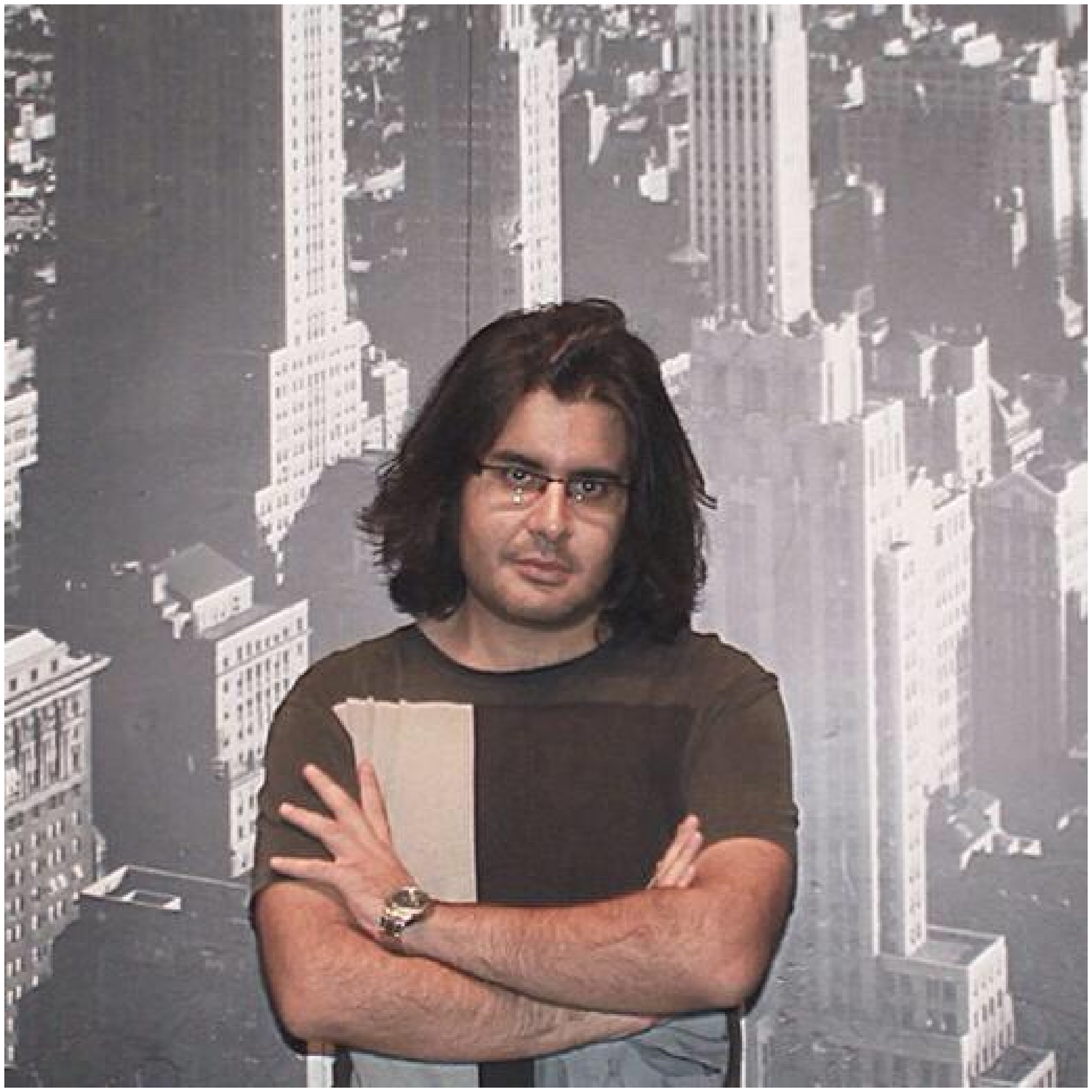}}]
	{Vahid Behzadan} is a Ph.D. candidate in the department of Computer Science at Kansas State University. He received his MS in Computer Science from University of Nevada, Reno, and a B.Eng. in Communications and Computer Systems from the University of Birmingham, UK. His research interests lie in the intersection of artificial intelligence, security, and complex adaptive systems. Contact him at behzadan@ksu.edu or 2184 Engineering Hall, 1701D Platt St., Manhattan, KS 66506.
\end{IEEEbiography}

\begin{IEEEbiography} [{\includegraphics[width=1.6in,height=1.35in,clip,keepaspectratio]{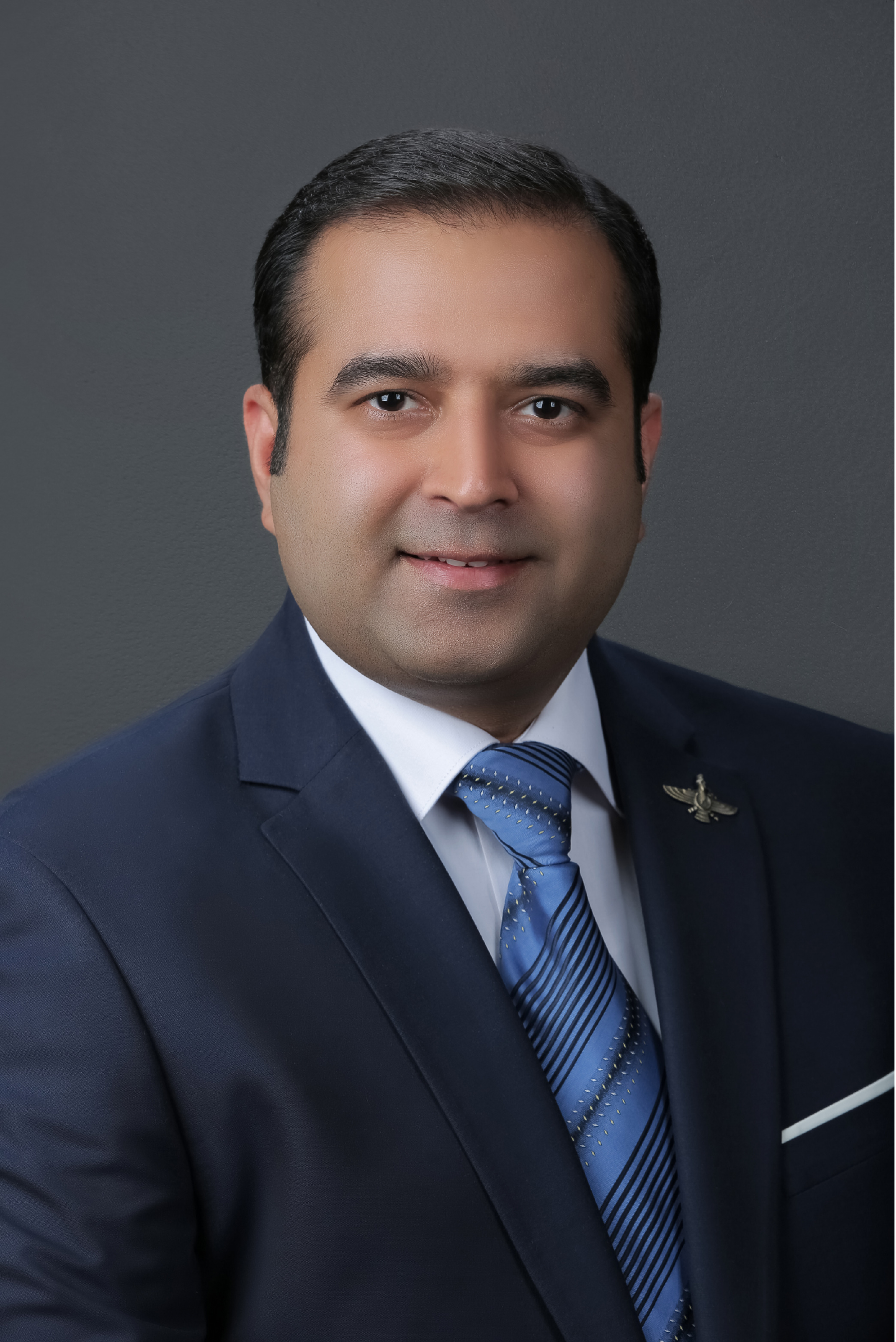}}]
{Arslan Munir} is currently an Assistant Professor in the Department of Computer Science at Kansas State University. He obtained his Ph.D. in Electrical and Computer Engineering from the University of Florida, Gainesville, USA. His current research interests include embedded and cyber-physical systems, secure and trustworthy systems, computer architecture, parallel and distributed computing, and artificial intelligence safety and security. He is a Senior Member of IEEE. Contact him at amunir@ksu.edu or 2162 Engineering Hall, 1701D Platt St. Manhattan, KS 66506.
\end{IEEEbiography}

\end{document}